\documentclass{article}

     \usepackage[nonatbib,preprint]{neurips_2020}

\usepackage[utf8]{inputenc} % allow utf-8 input
\usepackage[T1]{fontenc}    % use 8-bit T1 fonts
\usepackage{hyperref}       % hyperlinks
\usepackage{url}            % simple URL typesetting
\usepackage{booktabs}       % professional-quality tables
\usepackage{amsfonts}       % blackboard math symbols
\usepackage{nicefrac}       % compact symbols for 1/2, etc.
\usepackage{microtype}      % microtypography

\usepackage{xspace,amsmath, amssymb}

\usepackage{bm}
\usepackage{bbm}
\usepackage{graphicx}
\usepackage{subcaption}
\usepackage{xcolor}

\usepackage{tipa}

\def\bfu{{{\mathbf u}}}
\def\bfv{{{\mathbf v}}}

\def\bfx{{{\mathbf x}}}

\def\bfy{{{\mathbf y}}}
\def\bfz{{{\mathbf z}}}

\def\E{\ensuremath{{\mathbb E}}\xspace}

\def\R{\ensuremath{{\mathbb R}}\xspace}

\def\X{\ensuremath{{\mathcal X}}\xspace}

\DeclareMathOperator*{\argmin}{argmin}

\def\bfy{\mathbf y}

\def\bfo{\mathbf o}

\def\bfm{\mathbf m}

\def\learner{{\mathcal{A}}}
\def\vocab{V}
\def\pool{U}
\def\eval{E}

\title{Learning to Read through Machine Teaching}

\author{%
  Ayon Sen \\
  Department of Computer Science\\
  University of Wisconsin-Madison\\
  Madison, WI 53706 \\
  \texttt{asen6@wisc.edu} \\
  \And
  Christopher R. Cox\\
  Department of Psychology\\
  Louisiana State University\\
  Baton Rouge, LA 70803\\
  \texttt{chriscox@lsu.edu} \\
  \And
  Matthew Cooper Borkenhagen\\
  Department of Psychology\\
  University of Wisconsin-Madison\\
  Madison, WI 53706 \\
  \texttt{cooperborken@wisc.edu} \\
  \And
  Mark S. Seidenberg\\
  Department of Psychology\\
  University of Wisconsin-Madison\\
  Madison, WI 53706 \\
  \texttt{seidenberg@wisc.edu} \\
  \And
  Xiaojin Zhu\\
  Department of Computer Science\\
  University of Wisconsin-Madison\\
  Madison, WI 53706 \\
  \texttt{jerryzhu@cs.wisc.edu} \\
}

\begin{document}

\maketitle

\begin{abstract}

Learning to read words aloud is a major step towards becoming a reader.
Many children struggle with the task because of the inconsistencies of English spelling-sound correspondences. 
Curricula vary enormously in how these patterns are taught.
Children are nonetheless expected to master the system in limited time (by grade 4). 
We used a cognitively interesting neural network architecture to examine whether the sequence of learning trials could be structured to facilitate learning.
This is a hard combinatorial optimization problem even for a modest number of learning trials (e.g., 10K). 
We show how this sequence optimization problem can be posed as optimizing over a \emph{time varying distribution} i.e., defining probability distributions over words at different steps in training.
We then use stochastic gradient descent to find an optimal time-varying distribution and a corresponding optimal training sequence.
We observed significant improvement on generalization accuracy compared to baseline conditions (random sequences; sequences biased by word frequency).
These findings suggest an approach to improving learning outcomes in domains where performance depends on ability to generalize beyond limited training experience.

\end{abstract}

\section{Introduction}
Experiences unfold through time, and learning happens along the way. How learning events are sequenced has an important impact on knowledge acquisition.
Educational curricula incorporate sequential structure at multiple scales such as the organization of a single class, the arc of a semester, or the trajectory of a degree-granting program.
Our research addresses whether machine learning outcomes can be improved by optimizing the sequence of learning experiences in a complex knowledge domain.
Gains in the efficiency of learning could mitigate limitations on human learning arising from perception, attention, memory, and other aspects of human cognition \cite{Bengio2009, reisberg2013oxford}.

The production of serially ordered behavior has been studied since Lashley's \cite{lashley1951problem} classic work (see, e.g., \cite{botvinick2004doing}).
% Omit catastrophic interference note?
%However, how the order of experience influences learning has received less attention.
%Perhaps the most important finding was a negative one: \cite{mccloskey1989catastrophic} observed that simple neural networks exhibited ``catastrophic interference'' when the sequence of learning trials was blocked. People also exhibit unlearning under similarly restricted conditions {\color{red}(Ratcliff)}.
%This type of retroactive interference is less of a problem for people because different types of learning experiences are typically interleaved rather than blocked. Interleaving of experiences mitigates the problem of catastrophic interference in studies of human learning \cite{Carvalho2014}, connectionist networks \cite{hetherington1989catastrophic, mcrae1993catastrophic} and machine learning \cite{Li2012, Rau2013}.
%\cite{Bengio2009, Krueger2009, zhu2015machine}. 
Interleaving is a simple example of how ordering of learning experiences affects learning outcomes \cite{flesch2018comparing,kornell-2008-learning,kumaran2016learning,mccloskey1989catastrophic,Rau2013,rohrer-2007-shuffling}.
In other cases, the sequence is determined by structure of the to-be-learned material.
In elementary mathematics, for example, relations among addition, multiplication, and division dictate the order in which they are taught, allowing instruction to emphasize how one concept or operation participates in understanding the next, more sophisticated one.
Other work has suggested that starting simple and increasing problem complexity over time can help people learn more quickly while emphasizing structure that supports generalization \cite{Bengio2009}.
Benefits have been demonstrated in a variety of learning paradigms, including computational experiments involving shape recognition and other perceptual tasks, as well as language \cite{Elman1993, Newport1990, Rohde1999}.
	
Our research examined whether the sequence of learning trials could be optimized in an artificial neural network trained on a complex problem: learning the correspondences between the written and spoken forms of words in English.
Mastering these correspondences is an important step in becoming a skilled reader.
The material is difficult to teach and there is little agreement about how to do it.
Many children struggle at this early stage, with negative downstream effects on literacy and life experiences \cite{Seidenberg2017}.
Spelling-sound correspondences in English are systematic (letters and letter combinations represent sounds) but inconsistent (numerous forms deviate from central tendencies).
This quasiregular structure \cite{Seidenberg1989} exists at most levels of language \cite{Bybee2005}.
The spelling-sound system does not exhibit any obvious internal structure on which to base a learning sequence.
The system consists of numerous patterns differing in unit size, frequency, and consistency across words.
The question then is whether the sequence of learning experiences (training trials) can be ordered in a manner that facilitates acquiring this foundational knowledge.

In this paper, we approach sequence optimization as a gradient based optimal control problem for machine learning (i.e., machine teaching \cite{zhu2018overview}).
Given a learning algorithm, a pool of examples to choose from, and a target model to train, a machine teacher seeks to design a curriculum that conveys the solution efficiently \cite{zhu2013machine}.
However, most of the problems tackled so far in this domain work under a batch setting \cite{zhu2013machine,liu2016teaching} or for short training sequences \cite{sen2018machine}.

Our objective is to discover a training sequence such that the generalization performance of the learner is maximized given a fixed sequence length of 10,000 words.
Critically, this is too short to establish reading proficiency---a critical constraint analogous to the time pressure experienced by children.
Yet, the sequence is long enough to represent a useful amount of experience in early development and to pose a hard combinatorial problem, given that we explore sequences with as many as 1,000 unique words in any order.
A sequence of this length represents the quantity of words a child would experience from reading $\approx 80$ books appropriate to an early reader. % (assuming 125 words per book on average).

The main contributions of this paper are as follows. 
First, we formalize the problem of learning to read aloud as an optimal control problem.
We then show how a teacher can solve this problem in two steps: 1) using stochastic gradient descent to find optimal distribution of the words at different steps (\emph{time varying distribution}) of the sequence, 2) sampling an optimal sequence from the optimal time varying distribution.
Finally, we experiment with two ecologically valid vocabularies composed of either words a child or an adult is likely to encounter.
Our results show significant improvement over training sequences sampled using metrics that reflect the prevalence of words in a child or adult's language environment.

\section{Related Work} \label{sec:relwork}
\subsection{Optimal Control for Machine Learning}
Optimal control for machine learning (i.e., machine teaching) has been studied in many settings including cognitive psychology and education \cite{zhu2018overview}.
%Previous work has focused on optimizing the training set (i.e., batch) rather than the training sequence \cite{patil2014optimal,singla2014near}. 
%
%\cite{zhu2013machine,liu2016teaching}.
%It was first shown in \cite{lessard2018optimal} how machine teaching can be formulated as an optimal control problem.
%It has also been applied in the domain of machine learning security %\cite{liu2019unified,mei2015using} and human computer interaction %\cite{meek2016analysis}.
%
%
Patil et al. \cite{patil2014optimal} demonstrated that ideal training sets (i.e., batches) can be discovered for machine learners, and that different batches are ideal for learners with different constraints imposed.
Singla et al. \cite{singla2014near} trained human participants on visual classification problems with a machine teacher that greedily maximizes a submodular surrogate objective function to select examples from an ideal set.
The participants were modeled as selecting among decision rules stochastically, guided by feedback on their accuracy.
Participants trained by the machine teacher were significantly more accurate at classifying subsequent untrained images.
This work was extended in \cite{mac2018teaching} to explore the utility of additional instructive feedback.
%In \cite{cakmak2014eliciting}, the authors explore the scenario where a human teacher is responsible for training an agent by showing examples. 
%They study how to improve the teacher by giving teaching guidance.

An iterative version of machine teaching is explored in \cite{liu2017iterative}.
Here, the student is modeled as a linear learner, and the teacher can observe the characteristics of this learner during different training rounds and choose the next training example appropriately.
%To our knowledge, the only prior work to directly optimize the training sequence used hill climb search over a short sequence ($T=60$)\cite{rau2019using,sen2018machine}. 
Sen et al. \cite{sen2018machine}, and Rau, Sen and Zhu \cite{rau2019using} directly optimize the training sequence using hill climb search over a short sequence ($T=60$).
Sequences could be optimized to efficiently train an artificial neural network to identify chemistry molecule representations, and the sequences discovered by the search were more effective than those designed by human teachers.
% The authors use a hill climb search algorithm to find an optimal training sequence that performs better than a sequence designed by a human teacher. 

\subsection{Reading Development}
Early reading development rests on the ability of the child to learn as much as possible about how print and speech are related in order to move on to the more important, subsequent aspects of reading, namely comprehension \cite{Seidenberg2017}.
This creates an important issue for education: how can early reading experiences be structured to support speedy development, including the capacity to generalize \cite{Cox2019}?
The prevailing theory embodied in commercial reading curricula assumes that learners need to start with the smallest words that contain the most predictable spoken patterns, growing over time to be taught about longer words and larger, less predictable patterns of print.

Programs of instruction vary in their approach, in part because no comprehensive theory of how to structure the learner's print environment over time exists.
There are a variety of theories concerning which aspects of sublexical structure children learn to read, and in what order \cite{Vousden2011}, though no theory exists in this domain that deals with the corresponding teaching problem.
This is not suprising given its combinatorial complexity.
Making optimal choices about what to introduce into the child's experience and when requires experimentation over high dimensional aspects of language structure and how that structure can be introduced over time.
%The experimentation reported on here adds to this literature by casting early reading development as an optimization problem.
%Our approach seeks to understand how the learner's experience can be structured in such a way to learn as much as possible in a limited timeframe so that the child can learn quickly, and generalize to many untaught forms as possible.

\section{Preliminaries} \label{sec:background}
\subsection{Vocabulary and Feature Vectors}
We represent a vocabulary of monosyllabic words with $\vocab$.
Each word in $\vocab$ has an orthographic input ($\bfo \in \{0,1\}^{260}$) and a phonological output ($\bfy \in \{0,1\}^{200}$) representation.
The input and output patterns can represent up to 10 letters and 8 phonemes, respectively.
Each letter is encoded as a 26 dimensional one-hot vector.
Each phoneme is encoded as a 25 dimensional vector of articulatory features that accommodate all English speech sounds.
Each phoneme is encoded as $\bfm \in \{0,1\}^{25}$, such that each articulatory feature is either set or not.

Words are encoded such that the first vowel always occurs in the fourth letter or phoneme position.
For example $\vocab_{\text{coals}} = \left(\texttt{\_\_coals\_\_\_},\texttt{\_\_kolz\_\_} \right)$.
Notice that \texttt{oa} maps to the single vowel phoneme \texttt{o}.
The fifth letter position was also reserved for vowel representation: $\vocab_{\text{duct}} = \left(\texttt{\_\_ca\_t\_\_\_\_},\texttt{\_\_k@t\_\_} \right)$.
Note that padding `\_' is denoted by zero vectors in both input and output.%$\bfo_{\texttt{\_}}=\{0\}^{26}$ and $\bfy_{\texttt{\_}}=\{0\}^{25}$.

%We split the vocabulary into a training pool, $U=\{(\bfo_i,\bfy_i)\}_{i=1}^k$ and an evaluation set $E$.
%Details regarding the vocabulary and the split are given in Section~\ref{sec:exp}.
%We only consider the examples (words) in $U$ when training the learning algorithm.
%We evaluate the performance of the learned model on $E$.
%Note that there is no overlap between $E$ and $U$ i.e., any example that appears in $E$ does not appear in $U$.
%We do so as we are interested in the generalization performance of the learner.

\begin{figure}[!htb]
	\centering
	\includegraphics[width=0.7\textwidth]{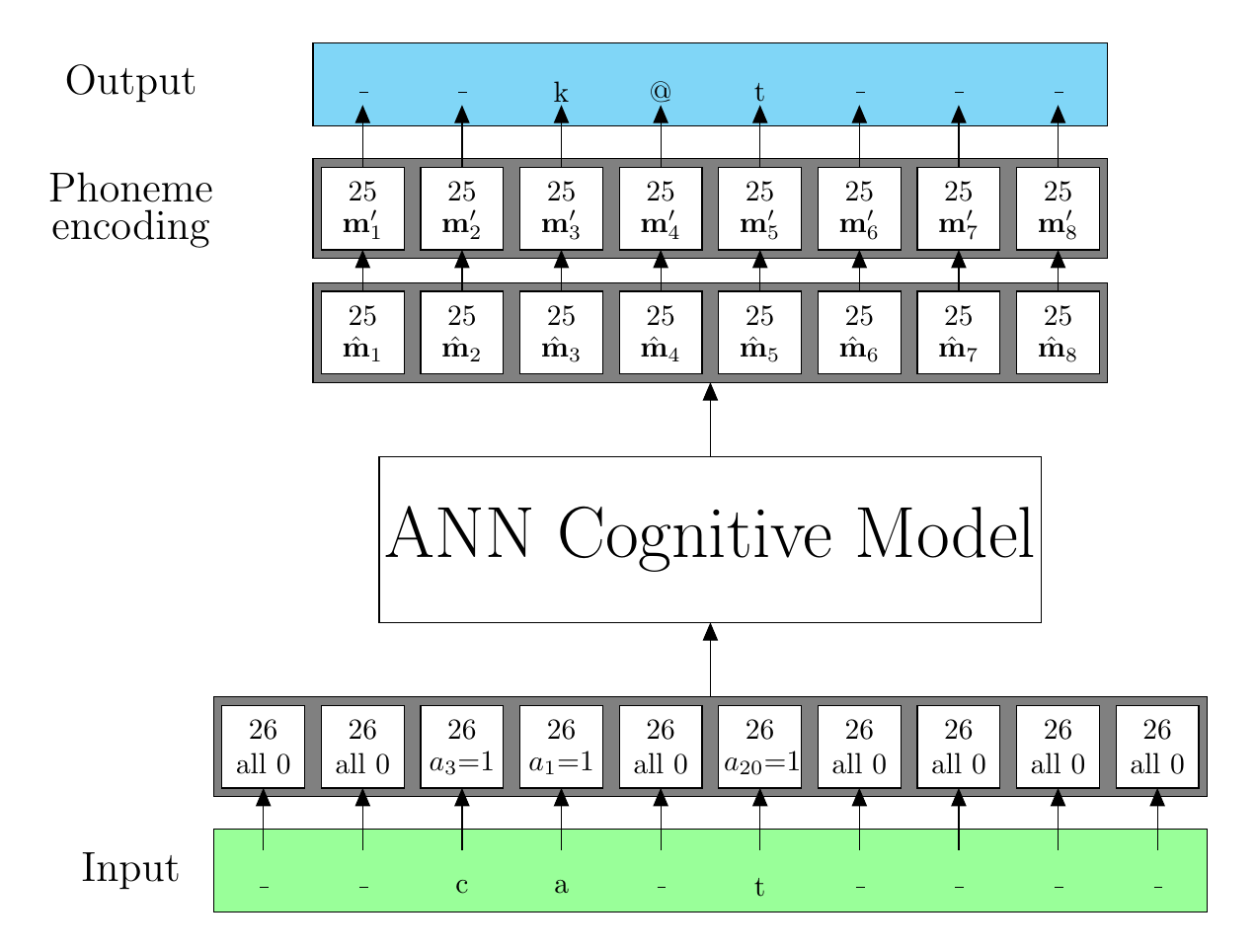}
	\caption{ANN cognitive model. It takes as input the orthographic representation of a word and predicts the phonological representation. $a_i$ indicates the $i$-th position of the input character vector.
	The continuous output vector is first decoded into individual phonemes, and then the complete phonological representation.}~\label{fig:ann}
\end{figure}

\subsection{Cognitive Model}
%We model learning to read using a feed forward artificial neural network (ANN).
We employ an ANN architecture with a long history of relevance in the cognitive science literature on reading development (Figure \ref{fig:ann}) \cite{Plaut1996, Seidenberg1989}.
It is a fully-connected feed-forward network with a single hidden layer (100 units) and sigmoid activation functions on all hidden and output units.
%Visual word recognition, including the development of the skill that underlies it, relies largely on the ability to associate the visual (orthographic) pattern of letters on the page to the corresponding speech sounds in the language.
%Learning to read can be accounted for %architecturally
%homogeneous architectures that gradually build up graded representations over repeated experiences of print and sound \cite{Harm2004, Plaut1996, Seidenberg1989}.
Our research builds on previous efforts to study teaching within a connectionist framework \cite{Harm2003}, specifically examining whether the sequence of learning trials can be optimized to support development.
The learner's environment consists of monosyllabic words that are presented one at a time, all letters simultaneously in parallel.
This allows for the use of a relatively simple, non-recurrent network while preserving essential aspects of early visual word recognition and its development, namely that a monosyllabic word can be taken in on a single visual fixation.

The learning procedure is defined by the dynamics $\bfx_{t+1} = f(\bfx_t, \bfu_t)$.
Here, $\bfx_t \in \X_t$ is the state of the model before training round $t$. In our setup, it is the vector of weights of the ANN.
The control input $\bfu_t$ is the orthographic, phonological representation pair for a word in the training pool: $\bfu_t \equiv (\bfo_t, \bfy_t) \in \pool \subset \vocab$.
This item is picked by a teacher and presented to the learner for training at round $t$.
The function $f$ defines the evolution of the state under external control.
Here, it is a backpropagation function with Nesterov momentum and cross entropy loss function.
The training round $t$ ranges from $0$ to $T-1$ where the time horizon $T$ is fixed at 10K.

After training is complete, the model can be used to make predictions denoted by $\hat{\bfy} = \learner(\bfx_T, \bfo)$.
Here $\hat{\bfy} \in [0,1]^{200}$.
We decode $\hat{\bfy}$ by first identifying individual phonemes.
Let $M$ denote the set of all possible phonemes (including padding).
Given a continuous vector $\hat{\bfm}_j \in [0,1]^{25}$, we decode it to a phoneme by $\bfm_j' = \argmin_{\bfm \in M} (\Vert\hat{\bfm}_j - \bfm \Vert_2)$.
%Given a continuous vector $\hat{\bfm}_j \in [0,1]^{25}$, we decode it to $\bfm'_j \in \{0,1\}^{25}$: $\bfm_j' = \argmin_{\bfm \in M} (\Vert\hat{\bfm}_j - \bfm \Vert_2)$.
Let $\bfy'\equiv [\bfm_1',\ldots,\bfm_8']$ i.e., $\bfy'$ denote the concatenation of the decoded phonemes.
Then the prediction is correct if $\bfy'=\bfy$.
We use the function $\rho:[0,1]^{200} \rightarrow \{0,1\}^{200}$ to denote the complete decoding procedure i.e., $\bfy' = \rho(\hat{\bfy})$.

\subsection{Teacher's Cost Functions}
The teacher takes into consideration two separate costs while designing a training sequence: a running cost and a terminal cost.
The running cost (denoted by $g_t(\bfx_t, \bfu_t)$) identifies how difficult/easy a problem is to teach to the learner.
The terminal cost on the other hand is defined on the final state of the learner i.e., the trained model:
\begin{equation}~\label{eq:terminal}
g_T(\bfx_T) = \frac{1}{|E|} \sum_{(\bfo, \bfy) \in E} \mathbbm{1}(\rho(\learner(\bfx_T,\bfo)) \ne \bfy)
\end{equation}
Here, $\eval=\vocab - \pool$ is a test set and $\mathbbm{1}(\cdot)$ is the indicator function.
The terminal cost estimates how well the learner generalizes to unseen examples.

\section{Optimal Control Problem} \label{sec:control}
In this section we define the teacher's optimal control problem and propose a solution.
The teacher's objective is to find a sequence of control inputs (training example sequence) that reduces the total cost.
%\begin{align}
%\min_{\bfu_0,\ldots,\bfu_{T-1}} &~~ g_T(\bfx_T) + \sum_{t=0}^{T-1}g_t(\bfx_t,\bfu_t) \nonumber\\
%s.t. &~~ \bfx_{t+1} = f(\bfx_t, \bfu_t), \bfu_t \in U, \forall t \nonumber\\
%&~~ \bfx_0 \text{ given} 
%\end{align}
We consider all examples to be equally difficult/easy. As the time horizon $T$ is fixed, total running cost is also fixed. 
This reduces the teacher's objective to
\begin{align}\label{eq:opt_prob}
\min_{\bfu_0,\ldots,\bfu_{T-1}} &~~ g_T(\bfx_T) \nonumber\\
s.t. &~~ \bfx_{t+1} = f(\bfx_t, \bfu_t), \bfu_t \in U, \forall t \nonumber\\
&~~ \bfx_0 \text{ given} 
\end{align}

We propose a gradient based solution to this problem in two steps.
We start by defining a \emph{time varying distribution} over the examples in $U$.
We assume the teacher has a \emph{start multinomial} $P = (p_1,\ldots,p_K)$ and an \emph{end multinomial} $Q = (q_1,\ldots,q_{K})$ over $U$.
Here, $K=|U|$.
At training round $t=0,\ldots,T$, the teacher uses an interpolated multinomial:
\begin{equation}
R_t = (T-t)/T*P + t/T*Q \label{eq_time_varying}
\end{equation}
The teacher will then draw $\bfu_t\sim R_t$ and train the learner with this example.
$(P,Q)$ then defines the time varying distribution.
We call this a time varying distribution as $R_t$ changes at each training round.
We find an optimal value for this pair: $(P^*,Q^*) = \argmin_{(P,Q)} E[\psi(P,Q)]$. 
Here $\psi(P,Q)$ denotes the terminal cost of a sequence drawn from the interpolated multinomials in (\ref{eq_time_varying}) defined by $(P,Q)$.
As $\psi(P,Q)$ is stochastic, we minimize over the expected terminal cost.
After identifying $(P^*,Q^*)$ we sample multiple sequences from the time varying distribution and pick the best one to solve (\ref{eq:opt_prob}).

This two step procedure has multiple advantages over solving (\ref{eq:opt_prob}) directly.
First, (\ref{eq:opt_prob}) is a combinatorial optimization problem. 
Such problems are hard to solve in practice, especially for long sequences.
Using a time varying distribution on the other hand allows us to optimize over a continuous space.
Moreover, directly solving the combinatorial optimization problem does not allow us to easily identify why a particular sequence is better than others.
But finding $(P^*,Q^*)$ allows us to readily understand different properties of good sequences.
For example, by inspecting $P^*,Q^*$ and $R^*_t$  we can identify examples that are more important during the various training rounds.

Because $(P, Q)$ are themselves multinomial distributions, they have a natural normalization constraint and are bounded. 
We circumvent this issue by reparametrizing $P$ as
$(\alpha_1,\ldots,\alpha_{K-1},\alpha_{K}=1) \in \R^{K}$, unconstrained.
Note that the last element is a constant $1$. We recover $P$ by $P_i = \exp \alpha_i / \sum_{j=1}^{K} \exp\alpha_i$.
Similarly $Q$ is reparametrized as $(\beta_1,\ldots,\beta_{K-1},\beta_{K}=1) \in \R^{K}$.
We now optimize $\bfz=(\alpha_1,\ldots,\alpha_{K-1},\beta_1,\ldots,\beta_{K-1}) \in \R^{2K-2}$: $\bfz^* = \argmin_{\bfz \in \R^{2K-2}} \E[\ell(\bfz)]$. $\E[\ell(\bfz)]$ denotes the expected value of the terminal cost function when training sequences are sampled using $\bfz$.
We find $\bfz^*$ by using stochastic gradient descent (sgd) with momentum
\begin{align}
\Gamma_{s + 1} = \gamma &\Gamma_{s} + \eta \nabla \E[\ell(\bfz_{s})] \nonumber \\
\bfz_{s + 1} &= \bfz_{s} - \Gamma_{s+1}\label{eq:sgd}
\end{align}
Here, $\eta$ and $\gamma$ are step size and momentum respectively. 
$\Gamma_0$ is set to 0. 
$\nabla E[\ell(\bfz_{s})]$ is the gradient of $E[\ell(\bfz_{s})]$ with respect to $\bfz_{s}$.
We estimate $\nabla E[\ell(\bfz_{s})]$ by using finite difference stochastic approximation\cite{flaxman2004online}:
\begin{equation}
\nabla \E[\ell(\bfz_{s})] \approx \E_{\bfv}[(\E [\ell(\bfz_{s} + \delta \bfv)] - \E [\ell(\bfz_{s})])\bfv] (2K-2) / \delta\label{eq:outerExp}
\end{equation}
Here, $\bfv \in \R^{2K-2}$ is a uniformly random unit vector and $\delta > 0$.
Note that the outer expectation is taken over $\bfv$.

\section{Experiments and Results} \label{sec:exp}
In this section, we empirically evaluate our proposed solution.
We compared our results against multiple baselines where the words are sampled with respect to their natural frequency in the environment.
We ran sgd with momentum in two steps.
First we found an optimal distribution ($\bar{P}^*$) from which the words can be drawn.
Note that this is not a time varying distribution.
Here, $\bar{P}_0$ was initialized as a uniform vector.
We ran sgd with momentum again to find the optimal time varying distribution by initializing $(P_0,Q_0)$ with $(\bar{P}^*, \bar{P}^*)$.

\begin{figure}[!htb]
	\centering
	\begin{subfigure}[t]{0.48\textwidth}
		\centering
		\includegraphics[width=\textwidth]{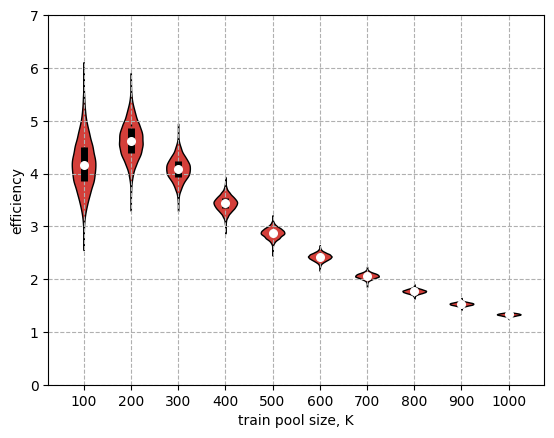}
		\caption{child}
	\end{subfigure}
	\begin{subfigure}[t]{0.48\textwidth}
		\centering
		\includegraphics[width=\textwidth]{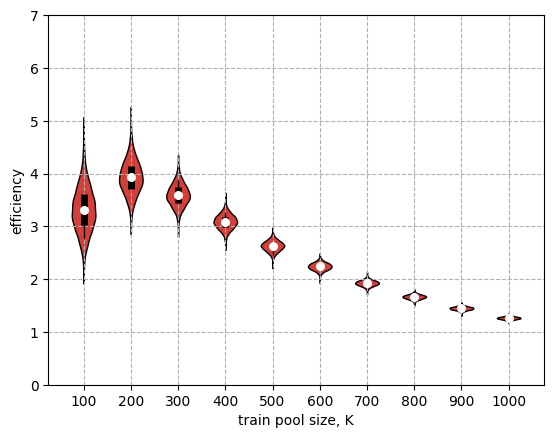}
		\caption{adult}
	\end{subfigure}
	\caption{Efficiency for different training pool sizes. Average efficiency peaks at training pools of size 200. The black bar shows the 25th and 75th percentile.}~\label{fig:efficiency}
\end{figure}

\subsection{Datasets}
We constructed two different training corpora of monosyllabic words, one using prevalence statistics relevant for adults and the other for children.
%Two different training corpora of monosyllabic words were constructed, one using prevalence statistics relevant for adults and the other for children.
Child words were selected if they appeared at least twice in the corpus of 250 children's books, and if a word possessed an age of acquisition (AOA) rating of 9 years old or younger from relevant norms \cite{Kuperman2012}.
Additionally, words needed to have a string length of greater than one and contain an orthographic vowel (e.g., the word "hmm" was discarded despite being included by the above criteria).
This yielded a vocabulary of 2869 words in total.
In order to have a corresponding set of words for comparison to the primary child set, we followed a similar procedure using prevalence statistics for adult texts.
The top 1000 most frequent monosyllabic words were selected from the Corpus of Contemporary American English (COCA) \cite{Davies2010} regardless of their presence in the child set.
An additional 1869 words were selected by their rank frequency past the 1000th most frequent monosyllabic word in COCA, skipping words that were present in the child set.
This sampling strategy was used in order to minimize the dependence of the two sets without skipping too many very common words.
This resulted in a comparison set with 947 words (34\%) also appearing in the child set.

Each vocabulary is divided into a training pool and test set.
First, we chose how many words should be a part of the training pool, evaluating training pools of different sizes.
For a particular size $K$, we randomly divided the vocabulary into a training pool and a test set.
Then we batch trained the ANN cognitive model on this training pool to convergence and calculated the number of words that are correctly predicted by the trained model in the test set.
We represent this value with $c$.
For this batch training we used a learning rate of $0.1$ for faster convergence.
We evaluated efficiency of size $K$ as $c / K$.
This process was repeated 100K times to find the average efficiency for a particular $K$.
We performed this experiment for $K=\{100, 200, \ldots, 1000\}$.
The results are shown in Figure \ref{fig:efficiency}.
It can be seen that on average $K=200$ is the most efficient and hence this particular $K$ value is used.
We also used $K=1000$ for further experiments as children are expected to encounter a larger variety of words in a classroom setting.
For a fixed $K$, we chose the training pool/test set split that resulted in the largest test set accuracy in the aforementioned experiment.

\begin{figure}[!htb]

	\centering
	\begin{subfigure}[t]{0.48\textwidth}
		\centering
		\includegraphics[width=\textwidth]{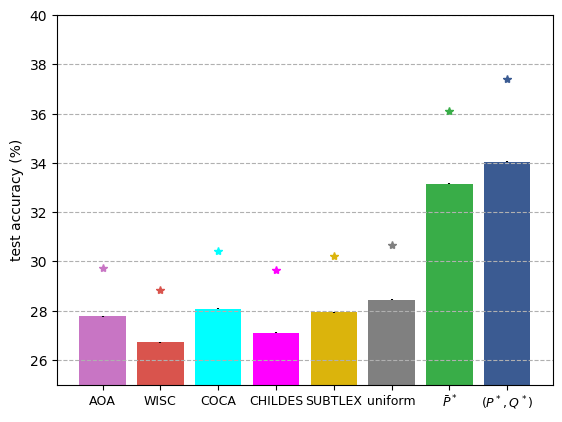}
		\caption{child, $K = 200$}
	\end{subfigure}
	\begin{subfigure}[t]{0.48\textwidth}
		\centering
		\includegraphics[width=\textwidth]{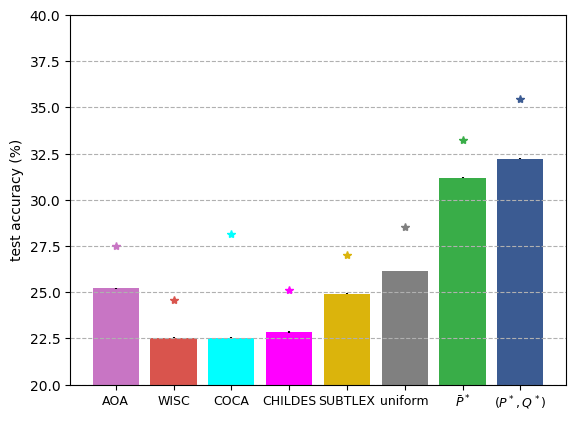}
		\caption{adult, $K = 200$}
	\end{subfigure}

	\begin{subfigure}[t]{0.48\textwidth}
		\centering
		\includegraphics[width=\textwidth]{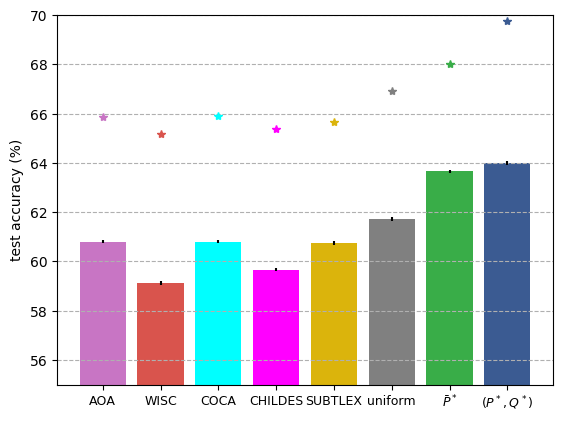}
		\caption{child, $K = 1000$}
	\end{subfigure}
	\begin{subfigure}[t]{0.48\textwidth}
		\centering
		\includegraphics[width=\textwidth]{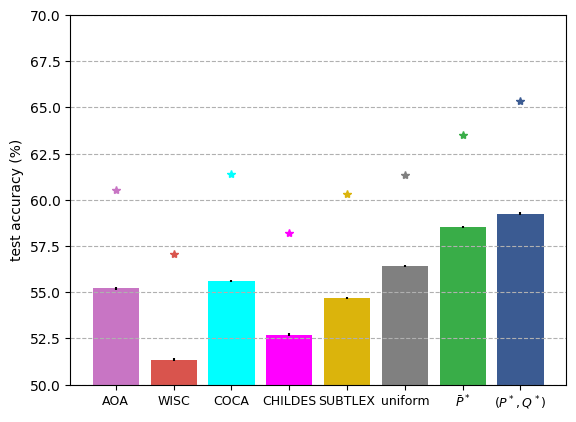}
		\caption{adult, $K = 1000$}
	\end{subfigure}
	\caption{Average test accuracy for different distributions sampled over 1000 sequences. The error bars show standard error. The optimal sequence test accuracy for each distribution is presented with an `*' above each bar.}~\label{fig:avg_test_accs}
\end{figure}

\subsection{Hyperparameters}
For the ANN cognitive model we used a learning rate of $0.02$ and a Nesterov momentum of $0.9$.
These values were motivated by previous explorations in this domain \cite{Cox2019}.
For sgd with momentum in (\ref{eq:sgd}), we set $\eta=0.01$ and $\gamma=0.9$.
We set $\delta=0.01$ to estimate $\nabla E[\ell(\bfz_{s})]$ in all experiments.
We approximated $\E_{\bfv}[(\E [\ell(\bfz_{s} + \delta \bfv)] - \E [\ell(\bfz_{s})])\bfv]$ in (\ref{eq:outerExp}) by sampling $20$ and $100$ uniform random unit vectors $\bfv$ for $K=200$ and $K=1000$ respectively.

\subsection{Baselines}
To compare our results we considered multiple baselines drawn from data that convey the prevalence of words in our experimental corpora in speech and print, many of which were also used for sampling words for inclusion in the corpora.
Given that printed and spoken language varies according to the audience for which it is designed, we gathered data for our words relevant to both adult and child audiences and from spoken and printed sources.
Child-directed speech data were taken from the CHILDES database \cite{MacWhinney2000}.
A baseline derived from word frequencies in child-directed text was also used, drawing from the same source that was used in sampling our child-directed words described previously (The Wisconsin Children's Book Corpus).
AOA data \cite{Kuperman2012} were used for estimates of the age at which words were learned from non-print based sources.
A baseline from adult-directed text frequencies utilized the COCA \cite{Davies2010}, a common large database used for text frequencies.
Paralleling the child-directed sources, values for child-directed speech were also used as a baseline, utilizing data from the SUBTLEXus databases \cite{Brysbaert2009}.
While SUBTLEXus isn't strictly a speech database, these data represent usage in movies, which we take to be speech-like.
Additionally, we implement a baseline using a uniform sampling probability over all candidate words, not favoring any particular word in the selection process.
The sampling probabilities drawn from these baselines serve as relevant comparisons for our optimized probabilities.

\subsection{Results}
%We now present the results of our experiments.
The expected test set accuracy of the optimal distributions, along with the baseline models, are presented in Figure \ref{fig:avg_test_accs}.
%Note that we also the present the results of the optimal distribution ($P^*$) here as well.
%This was found using the same procedure as $(P^*,Q^*)$.
These results show that we were able to find time varying distributions that are significantly better than the baselines.
The gains are higher for $K=200$ than $K=1000$.
It should also be noted that the overall test accuracy values found using $K=200$ are significantly lower than that of $K=1000$.
This is not surprising given that more irregular words are incorporated into the $K=1000$ training pool, meaning that they are not in the generalization set.
But in all cases the average test accuracy for $(P^*,Q^*)$ is statistically better than that of $\bar{P}^*$ and the other baselines ($p < 0.001$ for student's t-test).
The corresponding optimal sequence test set accuracy values are presented with `*' in the same Figure. %in Figure \ref{fig:best_test_accs}.
Not surprisingly, the optimal sequence found using $(P^*,Q^*)$ is always the best one.

In trying to determine those aspects of words that make them well suited to enhance learning, we calculated a number of word-level variables that might help us understand the distributions of $P^*$ and $Q^*$.
Orthographic neighborhoods were calculated as Levenshtein Distance with $D_{\text{Lev}} = 1$.
Phonological neighborhoods were determined by the number of words sharing the same orthographic body (i.e., the \emph{ushed} in \emph{hushed}) and phonological rime (i.e., the /\textipa{2}St/ in /h\textipa{2}St/ for \emph{hushed}).
For example, \emph{rushed} is a neighbor of \emph{hushed} but \emph{pushed} is not, despite sharing its body. Phonological density refers to the number of features on (i.e., equal to one) for a given word's target phonological representation.
Finally, three measures of Shannon Entropy were calculated, representing estimates of the predictability of a given orthographic unit's associated phonological code\cite{Siegelman2020}.
%\begin{equation}~\label{eq:entropy}
%[H(X) = -\sum p(X)\log p(X)]
%\end{equation}
The entropy of the oncleus unit (orthographic onset plus orthographic nucleus) is calculated with respect to any consonants that come before the orthographic vowel segment plus the vowel segment (i.e., the \emph{broo} in \emph{brook}).
Vowel entropy is calculated for the orthographic vowel segment (i.e., the \emph{oo} in \emph{brook}).
And the rime entropy is calculated for the orthographic vowel and everything that follows it (i.e., \emph{ook} in \emph{brook}).
Vowel entropy is calculated for the orthographic vowel segment (i.e., the \emph{oo} in \emph{brook}), and the rime entropy is calculated for the orthographic vowel and everything that follows it (i.e., \emph{ook} in \emph{brook}).
Three word prevalence measures are also included (AOA, child text frequency, and adult text frequency).
\begin{table}[!htb]
	\centering
	\caption{Correlations between word-level variables and mean of $P^*$ and $Q^*$. Correlations calculated as Spearman's $\rho$, and bolded if $p < .05$.}
	\resizebox{0.96\textwidth}{!}{%
		\begin{tabular}{l|r|r|r|r}
			Variable               & \multicolumn{1}{|c|}{Child, $K=200$} & \multicolumn{1}{|c|}{Adult, $K=200$} & \multicolumn{1}{|c|}{Child, $K=1000$} &\multicolumn{1}{|c}{Adult, $K=1000$} \\\hline
			Orthographic length    & 0.01      & \textbf{0.16}        & 0.02       & 0.01         \\
			Phonological length    & 0.1       & \textbf{0.23}        & 0.05       & 0.02         \\
			Orthographic neighbors & 0         & -0.1        & -0.05      & 0.02         \\
			Phonological neighbors & 0.02      & -0.08       & \textbf{-0.11}      & -0.03        \\
			Phonological density   & 0.11      & \textbf{0.15}        & 0.05       & 0.02         \\
			Morphology             & 0.05      & \textbf{-0.14}       & 0.05       & 0.02         \\
			Oncleus entropy        & 0.03      & \textbf{0.14}        & -0.01      & 0.02         \\
			Vowel entropy          & 0.07      & \textbf{0.22}        & 0.03       & 0.02         \\
			Rime entropy           & \textbf{-0.18}     & 0           & -0.05      & 0            \\
			Age of acquisition     & 0.05      & 0.11        & 0          & -0.01        \\
			Child text frequency   & -0.02     & -0.12       & 0.04       & 0.02         \\
			Adult text frequency   & 0.03      & -0.12       & -0.04      & 0
		\end{tabular}
	}
\end{table}

A few observations can be made about these correlational results in an attempt to understand what makes a word beneficial for learning.
Orthographic and phonological length correlate with average sampling probability in the adult, $K=200$ condition.
Some conditions tended to optimize for the predictability of subword orthographic units with respect to their phonology, namely the models trained on a candidate pool of 200 words.
However, the particular unit differs across conditions.
The child, $K=200$ condition is associated with higher sampling probabilites for rimes (word endings), and the corresponding adult condition with word-initial segments (onclei) and vowels.
This connects with findings in the reading development literature that have documented varying and sometimes conflicting findings about the location of statistical regularities that influence behavior in reading development \cite{Siegelman2020, Vousden2011}.
Little is explained by the structural properties included here for the conditions in with larger candidate pools. %which larger candidate pools were used.
Only the number of phonological neighbors seem to be related to mean of $P^*$ and $Q^*$ and only in the child model for the models trained with 1000 eligible words.

%\section{Discussion} \label{sec:disc}

\section{Conclusion} \label{sec:conc}

%Can the early development of knowledge about print be enhanced through intelligently selecting sequences of words for teaching?
%In quasiregular domains like that of the relationship between orthography and phonology in English, what is learned at a particular moment potentially bears on learning and generalization in subsequent reading experiences through partial regularities that exist across exemplars.
We have demonstrated a gradient based optimal control approach for discovering long sequences that achieves efficiency gains above and beyond batch optimization and frequency-weighted sampling.
Compared to the prior state of the art, which was restricted to applications with short sequences, successful optimization over $T=10,000$ with as many as $1,000$ unique elements is a noteworthy advance.

The learning environments employed in our simulations were constructed for their relevance to development as established from prevalence statistics for child and adult print and speech.
We find an optimal time varying distribution defined using two distributions over training words, which can be used to establish good training sequences for the learner.
Having done so in a cognitive architecture that simulates visual word recognition and development \cite{Seidenberg1989,Plaut1996}, this is potentially a first step towards practical applications in reading education \cite{Seidenberg2017}.

While variability in selection probabilities assigned to words across candidate pools indicate words that are useful for learning and performance, the structural properties that lead to their utility need to be studied further.
The distributions for $P^*$ and $Q^*$ across conditions aren't substantially different in the optimization attempts reported here, despite success in optimizing relative to several motivated comparison conditions including a $P^*$ only distribution.
In future work, more needs to be done to explore the effects of defining the time varying distribution using more than two distributions, positioned in ways throughout the sequence that may show advantages over the linear interpolation scheme employed here.
This includes experimentation with non-linear interpolation functions over the distributions to define $R_t$.

Nonetheless, our results show the possibility and potential of optimizing the sequence of words an early developing reader is exposed to in order to enhance learning, including generalization to untaught words.

\section*{Broader Impact}
Reading is a fundamental cognitive skill that impacts all aspects of short and long term human development starting early in life. 
Our work seeks to develop  solutions for improving literacy outcomes through the use of efficient and scalable computational methods that can be applied to reading development for early, emerging readers. 
As a result, our work is for the common good and is oriented towards ameliorating the challenges posed by learning in complex deep orthographies (like English) in an effort to counter the deleterious effects of illiteracy related to this property of the system as well as other factors.
We understand that \emph{one size does not fit all} and the sequence found by our current approach may not benefit all learners due to individual differences.
However, these are the first steps in taking into consideration variability across individual learners when defining sequences that support learning.
We do not believe that the data reported on here draw from any particular cognitive or other biases. Biases are only relevant insofar as they are reflected in the frequency of use represented in the prevalence data we draw from in our methods, and the use of these data is very common across psychological, educational, and computational fields.
Hypothetically, our work could be applied to training set poisoning in a way that could be harmful. 
While this is not the focus of our work, we acknowledge the possibility of the technology being misapplied in ways that might have negative consequences in applications where a nefarious user poisons a training sequence to fool a machine learning system.

\section*{Acknowledgment}
This work is supported in part by NSF 1545481, 1623605, 1704117, 1836978, the MADLab AF Center of Excellence FA9550-18-1-0166, the Vilas Trust, University of Wisconsin-Madison (MCB, MSS), Deinlein Language and Literacy Fund (MSS), and IES grant R305B150003 (MCB). 
This research was performed using the compute resources and assistance of the UW-Madison Center For High Throughput Computing (CHTC) in the Department of Computer Sciences. 
The CHTC is supported by UW-Madison, the Advanced Computing Initiative, the Wisconsin Alumni Research Foundation, the Wisconsin Institutes for Discovery, and the National Science Foundation, and is an active member of the Open Science Grid, which is supported by the National Science Foundation and the U.S. Department of Energy's Office of Science.
The opinions expressed are those of the authors and do not represent views of the US Department of Education.

\bibliography{biblio}
\bibliographystyle{plain}

\end{document}